\documentclass[10pt,english]{article}
\usepackage{graphicx}
\usepackage[T1]{fontenc}
\usepackage[latin9]{inputenc}
\usepackage[margin=1.1in]{geometry}
\usepackage{float}
\usepackage{amsmath}
\usepackage{amsbsy}
\usepackage{setspace}
\usepackage{amssymb}
\usepackage{esint}
\usepackage{cite}
\newfloat{algorithm}{tbp}{loa}
\floatname{algorithm}{Algorithm}
\usepackage{algorithmic}

\newlength\myindent
\makeatletter

\floatstyle{ruled}
\newfloat{algorithm}{tbp}{loa}
\floatname{algorithm}{Algorithm}

\usepackage{babel}

\begin{document}

\title{Reservoir Computing on the Hypersphere}

\author{M. Andrecut}

\date{April 24, 2017}

\maketitle
{

\centering Calgary, Alberta, T3G 5Y8, Canada

\centering mircea.andrecut@gmail.com

} 
\bigskip 
\begin{abstract}
Reservoir Computing (RC) refers to a Recurrent Neural Networks (RNNs) framework, 
frequently used for sequence learning and time series prediction. The RC system 
consists of a random fixed-weight RNN (the input-hidden reservoir layer) and a 
classifier (the hidden-output readout layer). Here we focus on the sequence 
learning problem, and we explore a different approach to RC. More specifically, 
we remove the non-linear neural activation function, and we consider an orthogonal 
reservoir acting on normalized states on the unit hypersphere. Surprisingly, our 
numerical results show that the system's memory capacity exceeds the dimensionality of 
the reservoir, which is the upper bound for the typical RC approach based on Echo 
State Networks (ESNs). We also show how the proposed system can be applied to 
symmetric cryptography problems, and we include a numerical implementation.
\smallskip


\smallskip

\end{abstract}

\section{Introduction}
Recurrent Neural Networks (RNNs) are a class of machine learning methods, frequently 
used to solve complex temporal tasks like sequence learning and time series prediction. 
The main characteristic of RNNs is their feed-back connectivity, which transforms them 
into complex information processing systems that can approximate any other non-linear 
dynamical systems with arbitrary precision (universal approximation theorem) \cite{key-1,key-2}. 
The generic dynamical equations describing the RNN state at time $t$ are given by:
\begin{align}
x(t) &= f(Vx(t-1) + Us(t)),\\
y(t) &= g(Wx(t)),
\end{align}
where $s(t) \in \mathbb{R}^M$, $x(t) \in \mathbb{R}^N$, and $y(t) \in \mathbb{R}^K$ are 
the input, internal (hidden) and respectively output units. The matrices $U \in \mathbb{R}^{N \times M }$, 
$V \in \mathbb{R}^{N \times N }$, and $W \in \mathbb{R}^{K \times N}$ are the input-hidden, 
hidden-hidden, and respectively hidden-output weights of the connections. The activation 
and the output functions, $f:\mathbb{R}^N \rightarrow \mathbb{R}^N$ and $g:\mathbb{R}^N \rightarrow \mathbb{R}^K$, 
are the non-linear sigmoid ($\tanh $) and respectively softmax functions. 

Training the RNN involves adjusting the connectivity weights in order 
to implement the desired temporally dependent input-output mapping $x(t)\rightarrow y(t)$. 
While the complexity of the RNN architecture greatly increases the processing capability 
and significantly extends the range of potential applications, it also requires difficult 
training procedures, such as error backpropagation which is computationally expensive, 
it has a slow convergence and often leads to suboptimal solutions \cite{key-3}.

A different approach, avoiding these difficulties, in based on the Reservoir Computing (RC) 
framework (see Ref. \cite{key-4} for a review). In this approach only the hidden-output weights $W$ are modified, leaving the rest 
of the weights $V$, $U$ unchanged and therefore greatly simplifying the learning process. 
Thus, the RC system consists of a random fixed-weight RNN (the input-hidden reservoir layer $f$,$V$,$U$) 
and a classifier (the hidden-output readout layer $g$,$W$). The fundamental property of RC is the 
intrinsic memory effect, due to the recurrent connections in the reservoir $V$. 

RC comprises several RNN models, including the popular Echo State Network 
(ESN) which is also the source of inspiration for the model discussed here \cite{key-5}.
ESN requires the contractivity of the hidden reservoir state transition function ($f,V,U$), which ensures 
the stability of the network dynamics, such that the effect of any given input perturbation $s(t)$ on the 
hidden state $x(t)$ will vanish after a finite number of time steps \cite{key-6}. A widely accepted solution is to 
rescale the reservoir matrix $V$ such that its spectral radius satisfies $r(V) < 1$, and to use the $\tanh$  
function for the activation of the hidden states. Other practical aspects concerning the ESNs 
implementation involve setting the sparsity of the connectivity matrices $V,U$, and the scaling factor of the 
input matrix $U$ \cite{key-7}. The main characteristic of the ESN is its memory capacity $\mu$, which is  
its ability to retrieve the past information. It has been shown that the memory capacity of the ESN is 
bounded by the dimensionality of the reservoir matrix, $\mu \leq N$ \cite{key-8}. An important parameter influencing 
the memory capacity is the spectral radius of the reservoir matrix, such that the memory capacity approaches 
its maximum value when the spectral radius is closer to one \cite{key-9}.

Here, we focus on the sequence learning problem and we discuss a substantially different RC model. 
More specifically we remove the non-linear neural activation function, and we constrain the dynamics 
of the system to the unit hypersphere by using orthogonal reservoir matrices. 
Our numerical results show that surprisingly the system's memory capacity exceeds the dimensionality 
of the reservoir, which is the upper bound of the ESNs. We also discuss how the proposed system can be 
applied to associative sequence learning and symmetric cryptography problems. 
A numerical implementation is also included in the Appendix.

\section{RC on the hypersphere}

As mentioned in the introduction, here we focus on the sequence learning problem, and therefore we model 
the input $s(t)$ as a discrete random variable drawn from a set of $M$ distinct and orthogonal 
states $\mathcal{S} = \{0,1,...,M-1\}$, encoded by the columns of the identity matrix $I=[\delta_{i,j}]_{M \times M}$, where:
\begin{align}
 \delta_{i,j} &= 
  \begin{cases}
   1  & \text{for } i=j \\
   0 & \text{for } i\neq j
  \end{cases}.
\end{align}
and
\begin{equation}
s(t) = m(t) \Leftrightarrow s(t) = [\delta_{0,m(t)},...,\delta_{m(t),m(t)},...,\delta_{M-1,m(t)}]^T = [0,...,1,...0]^T.
\end{equation}
Similarly, the desired output $y(t)$ is also a sequence, and therefore we also model it as a discrete random 
variable drawn from a set of $K$ distinct and orthogonal states $\mathcal{Y} = \{0,1,...,K-1\}$, encoded by the columns of the identity 
matrix $I=[\delta_{i,j}]_{K \times K}$:
\begin{equation}
y(t) = k(t) \Leftrightarrow y(t) = [\delta_{0,k(t)},...,\delta_{k(t),k(t)},...,\delta_{M-1,k(t)}]^T = [0,...,1,...0]^T.
\end{equation}
Thus, both input and output variables satisfy the constraint $\Vert s(t) \Vert = \Vert y(t) \Vert = 1$. 

We also require that the scaling input matrix $U \in \mathbb{R}^{N \times M }$ ($M \leq N$) has unit length 
columns, such that $\Vert Us(t) \Vert = 1$ and $\langle Us(t) \rangle = 0$, $\forall s(t) \in \mathcal{S}$. This can be easily 
achieved by generating a matrix $U$ with the elements uniformly distributed in $[0,1]$ and then performing 
the following two steps normalization:
\begin{align}
u_m &\leftarrow u_m - \langle u_m \rangle, \\
u_m &\leftarrow u_m/\Vert u_m \Vert,
\end{align}
where $u_m$, $m=0,1,...,M-1$, are the columns of $U$.

Now we constrain the dynamics of the hidden state on the unit hypersphere $\Vert x(t)\Vert = 1$ by using random orthogonal reservoir matrices, 
which can be easily obtained for example by using the QR decomposition of any normal distributed random reservoir matrix $V=QR$, 
and replacing $V$ with the orthogonal factor $Q$. Since $Q$ is orthogonal, $Q^TQ=QQ^T=I$, it is also an isometry, and therefore we 
have $\Vert Q x(t)\Vert = \Vert x(t)\Vert = 1$, which means that the spectral radius is $r(Q)=1$.

Finally, we remove the non-linear neural activation function $f$, and we consider the following generic equations 
describing the dynamical system:
\begin{align}
x(t) &= \frac{Qx(t-1) + Us(t)}{\Vert Qx(t-1) + Us(t) \Vert},\\
y(t) &= g(Wx(t)),
\end{align}
with the output $g$ given by the softmax function:
\begin{equation}
y_k (t) = \frac{\exp(\langle w_k, x(t) \rangle)}{\sum_{i=0}^{K-1} \exp(\langle w_i, x(t) \rangle)}, \; k = 0,1,...,K-1,
\end{equation}
where $w_k$ is the row $k$ of the matrix $W$, and $\langle.,.\rangle$ is the standard dot product.
The learning of the $W$ weights can be performed either offline (batch) or online (iteratively). 

In the offline learning setting, one collects all the hidden states and the desired outputs as columns 
in the matrices $X=[x(t)]_{N \times T}$ and respectively $Y=[y(t)]_{K \times T}$, $t=0,1,...,T-1$, 
and computes the output weights matrix using \cite{key-6,key-7}:
\begin{equation}
W = YX^\dagger,
\end{equation}
where 
\begin{equation}
X^\dagger = \lim_{\eta \rightarrow 0_+}  X^T(XX^T + \eta I)^{-1}
\end{equation}
is the Moore-Penrose right pseudo-inverse of $X$.
\pagebreak 

For problems that require online learning, one can use 
iterative methods. For example one can see that the matrices $YX^T$ and $XX^T$ can be written iteratively 
as following:
\begin{align}
YX^T(\tau) = \sum_{t=0}^{\tau} y(t)x^T(t),\\
XX^T(\tau) = \sum_{t=0}^{\tau} x(t)x^T(t),
\end{align}
and therefore one can always find $W(\tau)$ for any time step $0 < \tau < T$. 
An alternative method, also frequently recommended in the literature, is the Recursive Least Squares (RLS) algorithm \cite{key-6,key-7}.

Unfortunately, both of these online methods are computationally expensive and suffer from numerical stability issues. 
Here we prefer to use a more simple method based on gradient descent, which also eliminates completely 
the need for matrix inversion. 

We observe that we can consider the learning of the output weights matrix $W$ as a classification problem with the 
training set $\{(x(t),y(t)) \; \vert \; y(t) \in \mathcal{Y}, t=0,1,...,T-1 \}$. Thus, we assume that:
\begin{equation}
p(k \equiv y(t)\vert x(t)) = \frac{\exp( \langle w_k, x(t) \rangle)}{\sum_{i=0}^{K-1} \exp(\langle w_i, x(t) \rangle))}.
\end{equation}
In order to learn the weights, we maximize the log likelihood function, or equivalently we minimize the cross entropy function:
\begin{equation}
\mathcal{H}(t) = - \sum_{k=0}^{K-1} y_k(t) \log p(k \vert x(t)).
\end{equation}
One can easily show that the gradient of $\mathcal{H}$ with respect to $w_k$ is:
\begin{equation}
\nabla_{w_k} \mathcal{H}(t) = (p(k \vert x(t)) - y_k(t)) x(t).
\end{equation}
Therefore, the online learning equation is given by:
\begin{equation}
W(t+1) = W(t) + (y(t) - p(t))x^T(t),
\end{equation}
where $W(0)=0$, and $p(t) \in \mathbb{R}^K$ is the vector with components:
\begin{equation}
p_k(t) \equiv p(k \vert x(t)), k=0,1,...,K-1.
\end{equation}

\section{Associative vs generative regimes}

As discussed so far the RC system works in an associative regime, which means that after learning one needs 
to feed the system with the sequence $s(t)$ in order to generate the associated output sequence $y(t)$. 
While this is the typical sequence to sequence learning scenario, another interesting possibility is the 
generative case where the system learns a sequence $s(t)$ and then it is able to generate the same 
sequence $s(t)$ starting from the initial state $s(0)$, and recursively feeding back the next predicted 
value $\hat{s}(t+1)$:

\begin{align}
x(t+1) &= \frac{Qx(t) + U\hat{s}(t)}{\Vert Qx(t) + U\hat{s}(t) \Vert},\\
\hat{s}(t+1) &= g(Wx(t+1)),
\end{align}

In this case we train the system starting from $x(0)=0$, and we replace $y(t)$ with $s(t+1)$, 
such that the online learning update equation becomes:
\begin{equation}
W(t+1) = W(t) + (s(t+1) - p(t+1))x(t+1)^T,
\end{equation}
where $p(t+1) \in \mathbb{R}^M$ is the vector with components:
\begin{equation}
p_m(t+1) \equiv p(m \equiv s(t+1) \vert x(t+1)), m=0,1,...,M-1.
\end{equation}
Also, for the offline learning case the matrix $Y$ is replaced with the matrix $S=[s(t)]_{M \times T}$, $t=0,...,T-1$,  
such that:
\begin{equation}
W = SX^\dagger.
\end{equation}

\section{Memory capacity}

In order to estimate the memory capacity of the system we consider the generative regime, and we define the memory capacity 
as the maximum length $T$ of the sequence that can be learned and reproduced with an error $0 \leq \varepsilon < \theta < 1$, 
given the size of the reservoir $N$ and the number of possible input states $M$. The error of the recalled sequence can 
be simply estimated as following:
\begin{equation}
\varepsilon = 1 - T^{-1} \sum_{t=0}^{T-1} \delta(s(t),\hat{s}(t)),
\end{equation}
where 
\begin{align}
 \delta(a,b) &= 
  \begin{cases}
   1  & \text{for } a=b \\
   0 & \text{for } a\neq b
  \end{cases}.
\end{align}

We expect that the memory capacity, and the recall error $\varepsilon$ to depend on both the size of the reservoir $N$ 
and the number of possible input states $M$. Therefore, it is convenient to fix the length $T$ of the learned sequences, 
and to define two quantities $\nu = N/T \in [0,1]$ and $\rho = M/T \in [0,1]$, 
and to estimate the two dimensional error function $\varepsilon=\varepsilon(\nu,\rho)$. 

\begin{figure}[!t]
\centering \includegraphics[width=15.2cm]{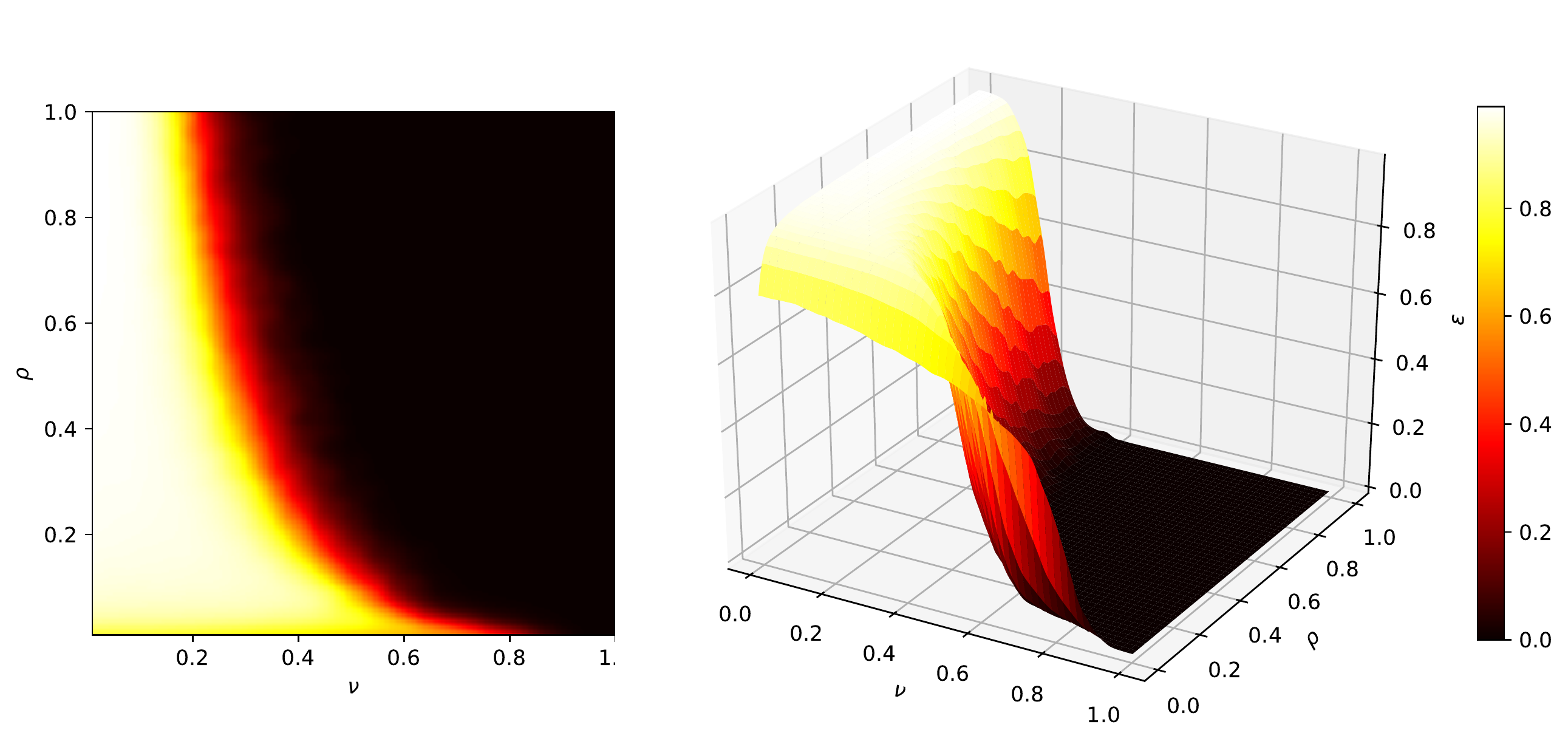}
\caption{The recall error $\varepsilon(\nu,\rho)$ as a function of 
$\nu = N/T \in [0,1]$ and $\rho = M/T \in [0,1]$.}
\vspace*{\floatsep}
\centering \includegraphics[width=15.2cm]{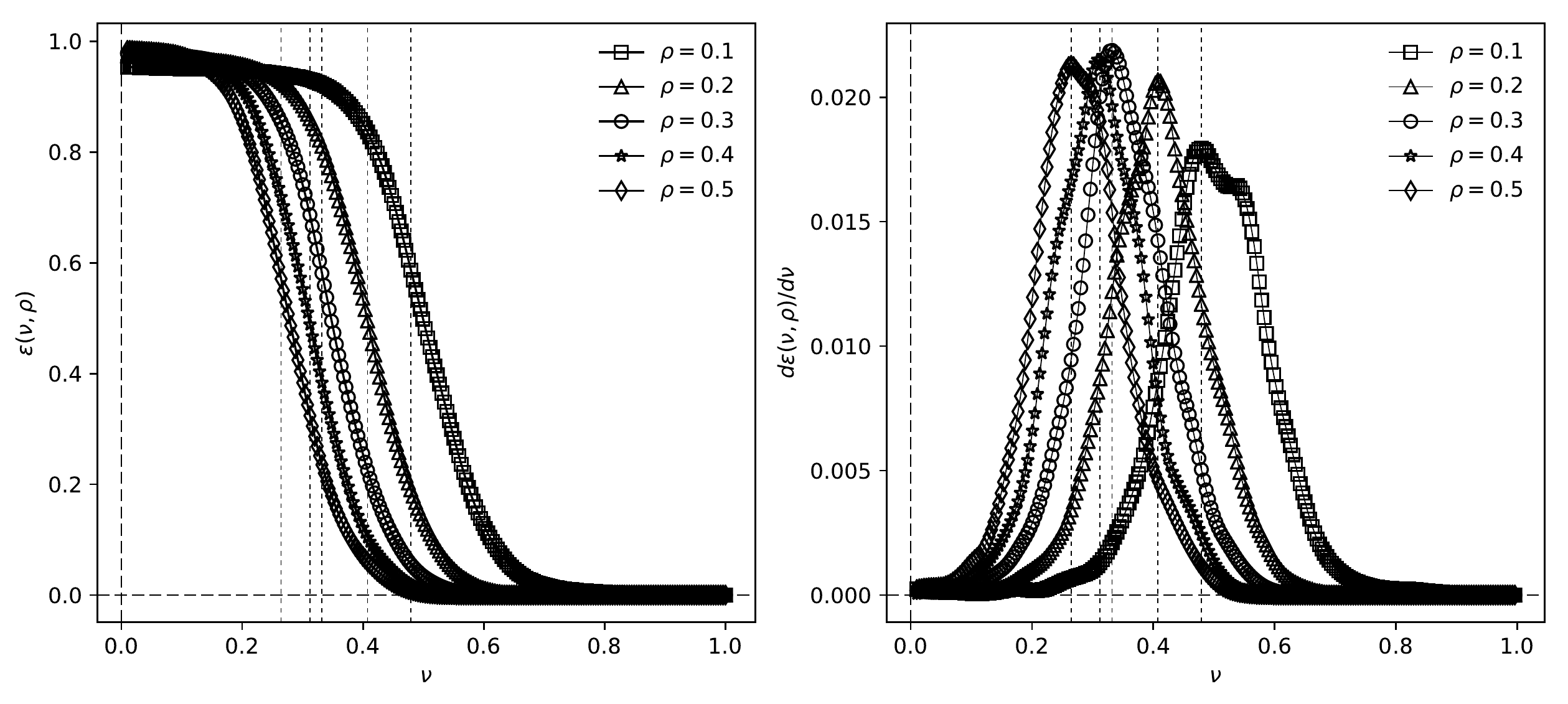}
\caption{The recall error $\varepsilon(\nu,\rho)$ (top) and the derivative 
$d\varepsilon(\nu,\rho)/d\nu$ (bottom) as a function of $\nu = N/T \in [0,1]$ and $\rho = M/T \in [0,1]$. 
The dotted vertical lines correspond to the transition points (the maximum of the derivative). 
}
\end{figure}

In Figures 1 and 2 we give the results obtained for $T=10^3$ averaged over $10^3$ random trials, 
using the pseudo-inverse learning. 
One can see that there is a wide range of the parameters $(\nu,\rho)$ where the recall is perfect, 
and in this region we have $N<T$, which means that the memory capacity exceeds the size of the reservoir. 
\pagebreak 

\begin{figure}[!t]
\centering \includegraphics[width=15.2cm]{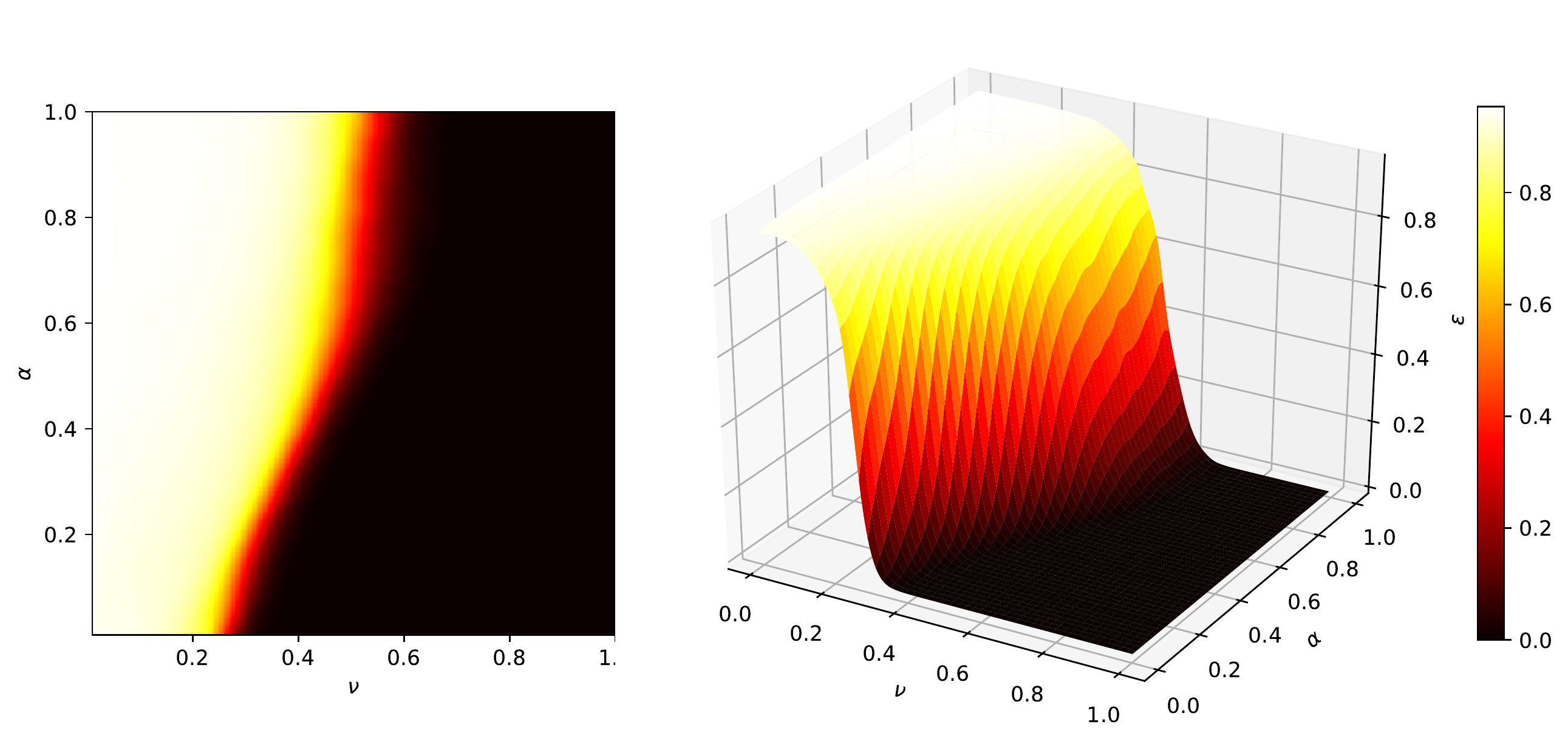}
\caption{The recall error $\varepsilon(\nu,\rho,\alpha)$ for a fixed $\rho=0.1$, 
as a function of $\nu = N/T \in [0,1]$ and the leaking integration rate $\alpha \in [0,1]$.}
\vspace*{\floatsep}
\centering \includegraphics[width=15.2cm]{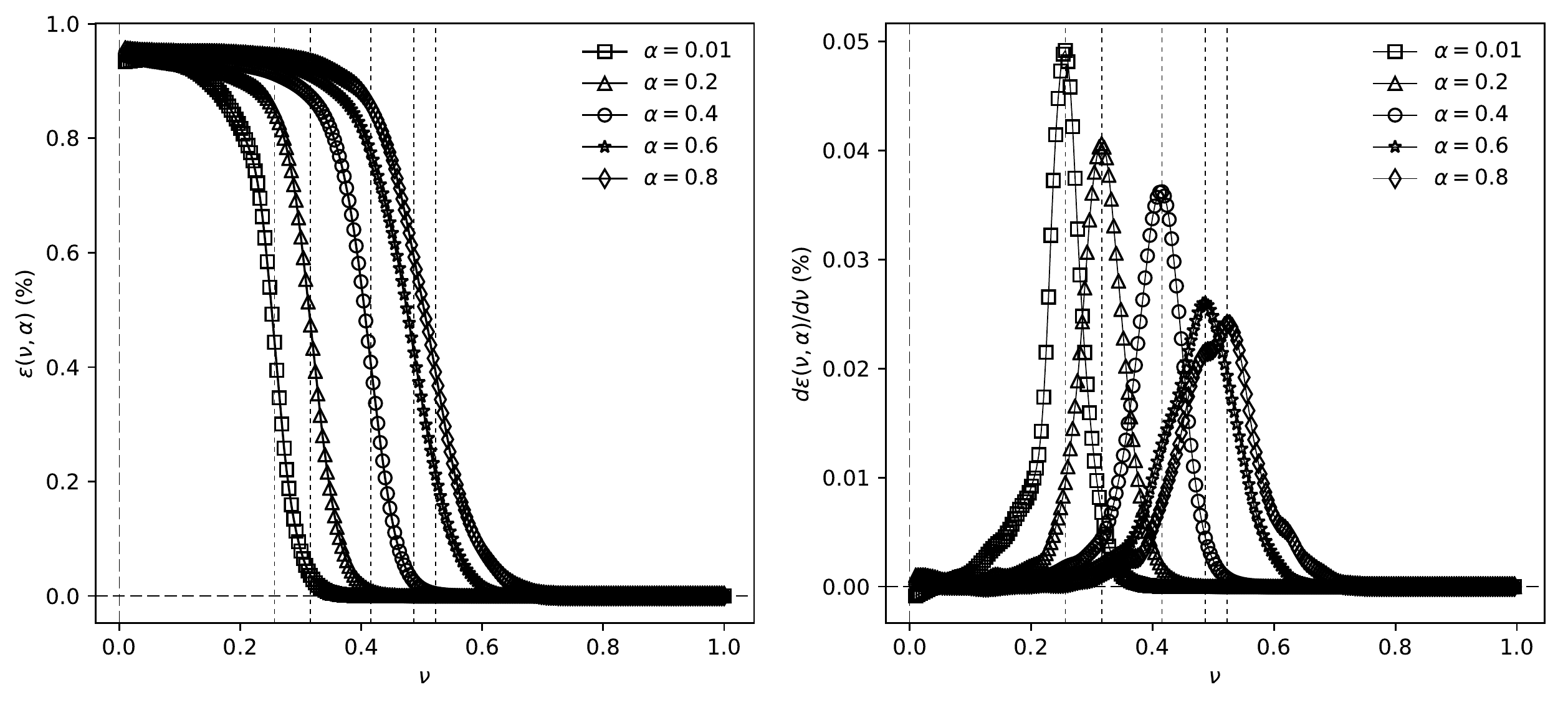}
\caption{The recall error $\varepsilon(\nu,\alpha)$ (top) and the derivative 
$\varepsilon(\nu,\alpha)/d\nu$ (bottom) for a fixed $\rho=0.1$, 
as a function of $\nu = N/T \in [0,1]$ and the leaking integration rate $\alpha \in [0,1]$.}
\end{figure}

We can further refine the model by considering the leaky-integration effect on the hidden state activation, 
as following:
\begin{align}
x(t+1) &= \frac{(1-\alpha )x(t) + \alpha (Qx(t) + U\hat{s}(t))}{\Vert (1-\alpha )x(t) + \alpha (Qx(t) + U\hat{s}(t)) \Vert},\\
\hat{s}(t+1) &= g(Wx(t+1)),
\end{align}
where $\alpha \in [0,1]$ is the integration rate, such that for $\alpha = 1$ we recover the initial basic model. 
For this model, the memory capacity, and therefore the recall error will depend on three 
parameters  $\varepsilon=\varepsilon(\nu,\rho,\alpha)$. 

In typical applications the number of distinct input states is much smaller than the length of the input sequence, 
$M\ll T$, and it is also fixed, like for example the alphabet in text learning. Therefore, in order to estimate 
the effect of leaky integration we consider a fixed value $\rho = M/T = 0.1$ and we vary the other parameters, 
$\nu \in [0,1]$ and $\alpha \in [0,1]$. 
The obtained results are shown in Figures 3 and 4. One can see that the leaky-integration procedure increases the 
memory capacity even more, such that the length of the sequences $T$ can be up to four times larger than the dimensionality 
of the reservoir $N$ when the leaking integration rate is small. 
Therefore, a third possibility is to set the integration rate $\alpha$ to a small value, $\alpha = 0.1$, and to vary $\nu \in [0,1]$ and $\rho \in [0,1]$. 
In this case, the numerical results are given in Figures 5 and 6, and one can see that the predominant effect is at low values 
of $\rho \le 0.2$, where the error is smaller, comparing to the case when $\alpha=1.0$, shown in Figures 1 and 2. 
However, for $\alpha>0.2$ the recall becomes slightly worse, increasing the fluctuations in the transition region. 
Thus, by adjusting the three parameters $(\nu,\rho,\alpha)$, one can fine tune the dynamics of the RC system in order to obtain the best recall values. 

\begin{figure}[!t]
\centering \includegraphics[width=15.2cm]{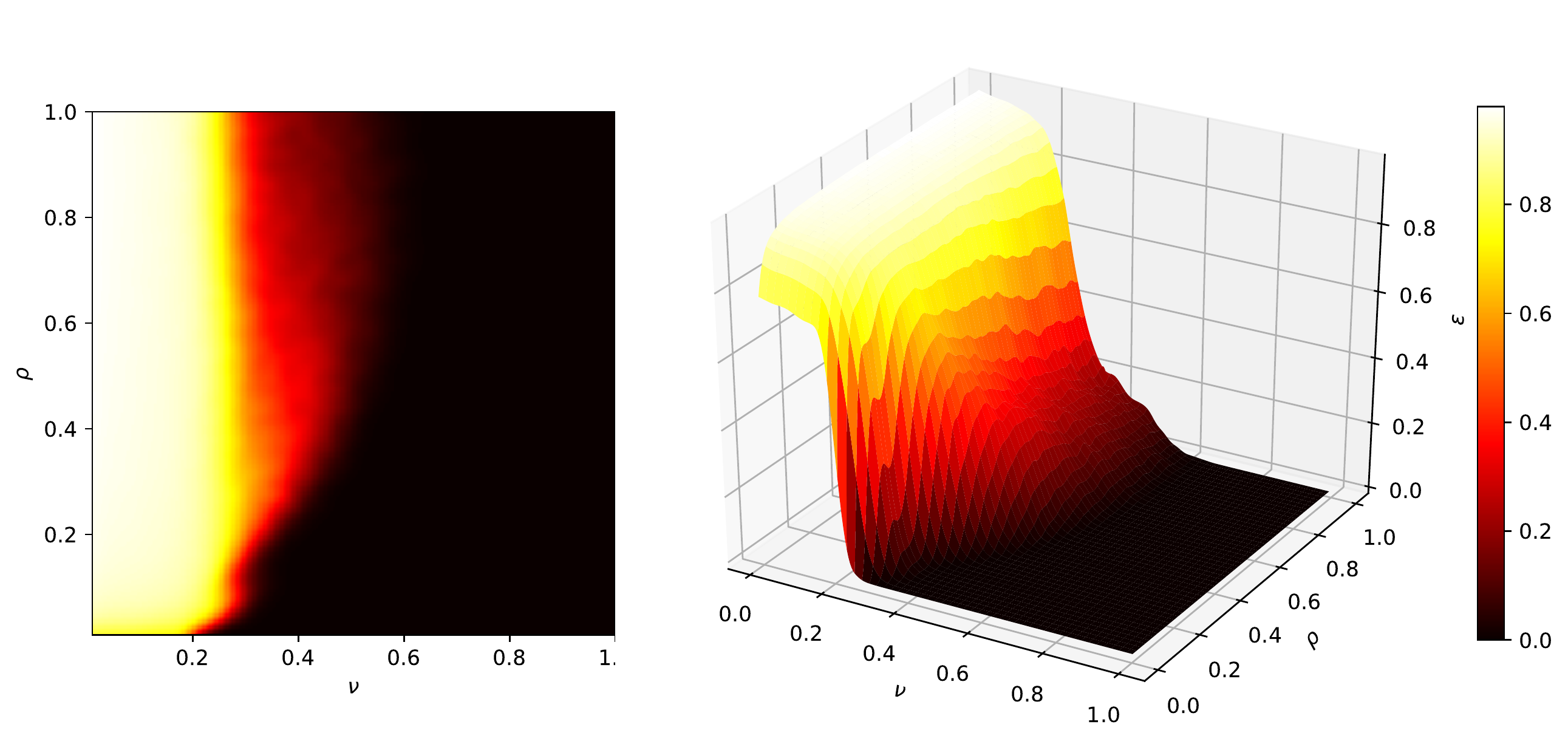}
\caption{The recall error $\varepsilon(\nu,\rho)$ as a function of 
$\nu = N/T \in [0,1]$ and $\rho = M/T \in [0,1]$ for a fixed integration rate $\alpha = 0.1$.}
\vspace*{\floatsep}
\centering \includegraphics[width=15.2cm]{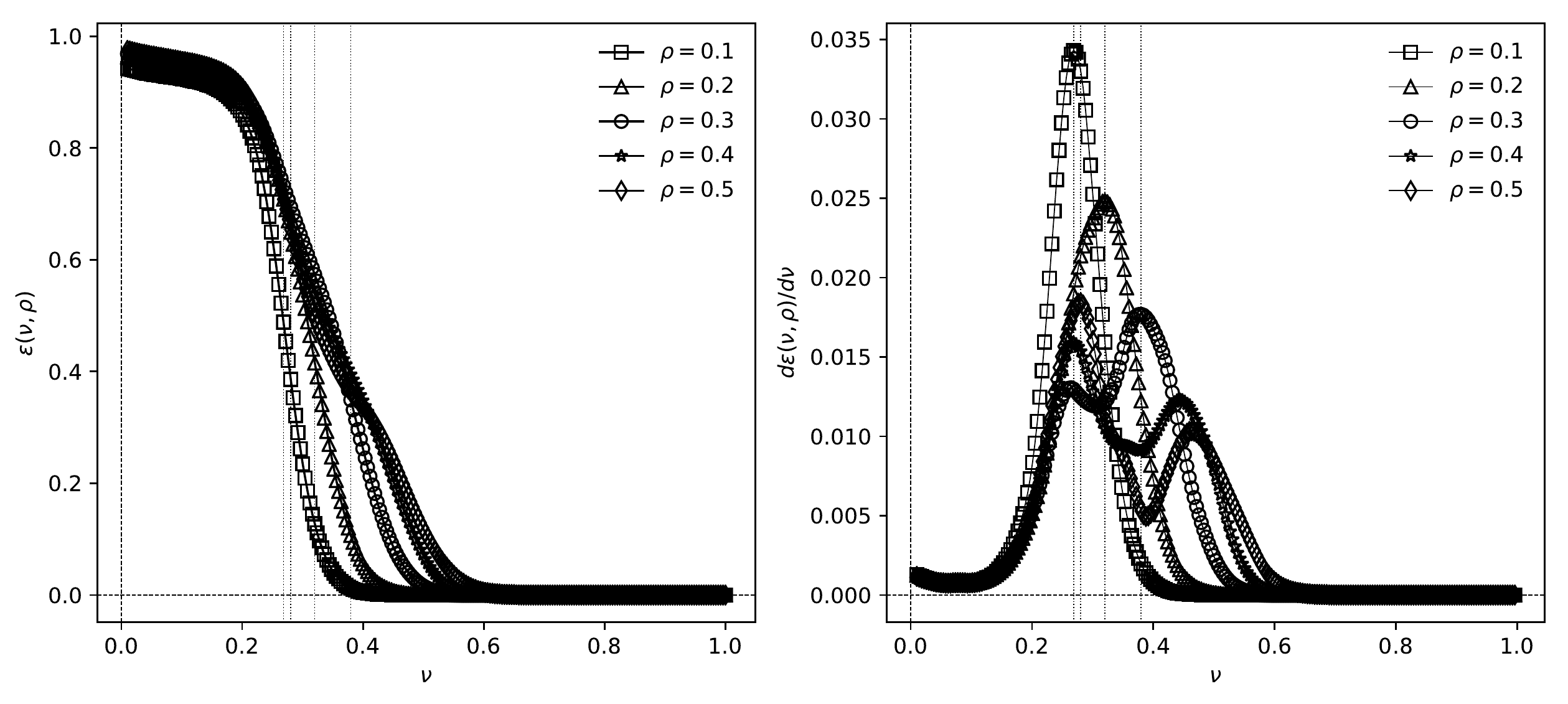}
\caption{The recall error $\varepsilon(\nu,\rho)$ (top) and the derivative 
$d\varepsilon(\nu,\rho)/d\nu$ (bottom) as a function of $\nu = N/T \in [0,1]$ and $\rho = M/T \in [0,1]$ for a fixed integration rate $\alpha = 0.1$.}
\end{figure}

The above numerical results show that the memory capacity of the proposed RC system exceeds the reservoir size, which is the upper bound of typical ESNs. 
We should note that in Ref. \cite{key-8} it is also mentioned that in certain conditions the ESNs are capable to 
exploit dependencies in the input data, in order to achieve a memory capacity exceeding the dimensionality of the 
reservoir. However, our results show that this is possible even without exploiting the dependencies (regularities) in the input data,  
and in fact this effect is mostly a consequence of the system's architecture. 
More specifically, an approximate solution for the weight matrix $W$ is still possible for $T > N$, via the pseudo-inverse or the gradient descent computation.  
The computation stability of the weight matrix $W$ is also improved by the orthogonal reservoir matrix $Q$, with a unit spectral radius $r(Q)=1$.  
The role of $Q$ is to optimally project the normalized hidden state vectors $x(t)$ on the hypersphere, which improves the conditioning of the matrix $X$, and therefore 
the stability of $W$. 
It turns out that such an approximation of $W$ is enough for the softmax classifier, which also doesn't require an exact solution.   
The softmax classifier computes the index corresponding to the maximum coordinate of $s(t+1) = g(Wx(t+1))$. 
Thus, an approximate output vector $s(t+1)$, obtained by projecting the hidden state $x(t+1)$ with an approximate matrix $W$, 
should be enough for the classification purpose, if it still has the maximum coordinate in the "right" place. 
The increased memory capacity of the discussed RC model is mainly a consequence of this relaxation of the solution.

As mentioned before, another parameter influencing the memory capacity is the number of possible distinct (orthogonal) input states $M$. 
If we look again at the results given in Fig. 1, we can see that an increase in $M$ reduces the error of the system, which is equivalent to an increase in the 
memory capacity.
This result is counter intuitive, since one would expect that it should be more difficult to learn complex sequences than simple ones. 
However, by increasing the number of distinct states $M$ we also increase the size $M \times T$ of the matrix $S\in \mathbb{R}^{M \times T}$, 
and implicitly the size $M \times N$ of the weight matrix $W = SX^\dagger \in \mathbb{R}^{M \times N}$, 
and therefore we increase the "physical support" of the memory. Thus, 
it is expected to see an increase in the memory capacity with an increase of $M$.

We have also noticed that the system's behavior is independent of the orthogonal matrix "type" used for the reservoir. 
With the QR method one obtains a full dense orthogonal matrix $Q$. However, one can use very sparse  
permutation matrices, which are also orthogonal, and the system maintains its memory capacity. 
In particular, the cyclic permutation matrices are very appealing since their application to the hidden state $x(t)$ consists in simply 
shifting all the elements one step to the front, and inserting the first element at the back. Therefore, 
the cyclic permutation mapping can be simply implemented as following:
\begin{equation}
x_n(t) \leftarrow x_{(n+1)\text{mod}N}(t), \; n=0,1,...,N-1,
\end{equation}
and it eliminates completely the need for the reservoir storage, and significantly speeds up the computation 
from $O(N^2)$ to $O(N)$.

\section{Symmetric cryptography application}

Recently it has been suggested that ESNs could also be used in cryptography \cite{key-10}. 
In this setting, Alice and Bob exchange messages and try to protect their communication 
against Eve's eavesdropping. 
In order to exchange messages, both Alice and Bob share an identical copy of an ESN. 
To encrypt a message, Alice trains the ESN such that the ESN reproduces the input. 
She then sends the output weights $W$ to Bob who uses them to decrypt the message.  
Without the corresponding ESN  internal structure ($U,V$) Eve will not be able to decipher the message. 
Here, obviously the internal structure of the ESN ($U,V$) is the 
secret key. 

The above approach, based on the generative regime of the RC (or ESN), is not the only solution. 
Here we describe a different method based on the associative regime, which also adds 
several extra security layers. Since we assume that the RC operates in the associative regime, 
in addition to the internal structure of the RC we can also use the input string $s(t)$ as a secret key ($U,Q,s(t)$), 
and consider that the output string $y(t)$ is the message to be encrypted. In order to increase the 
security we also use full dense orthogonal reservoir matrices $Q$.
In such a scenario both Alice and Bob have a copy of the RC. Alice uses the 
RC to encrypt a message $y(t)$ with the secret key ($U,Q,s(t)$), then she sends the output weights $W$ 
to Bob who uses them together with the same secret key ($U,Q,s(t)$) to decrypt the message. 
This, approach is also very convenient because the secret input string $s(t)$ can be 
randomly generated from a secret string (password), for example using a secure one-way 
hash function, such that blocks of text with the length equal to the hash function output 
can be encrypted and decrypted. 
Also, prior to the RC learning step, as an element of extra security, 
the message can be encrypted with a simple XOR step, using the same secret key $s(t)$: $y(t) \leftarrow y(t)\otimes s(t)$.
Obviously, the decryption also requires the application of this extra step. 
The XOR pre-encryption and post-decryption step can be used also for the generative regime, 
making it more robust to attacks. 

\section*{Conclusion}

In this paper we have presented a new RC model with the dynamics constrained to the  
unit hypersphere. The model removes the non-linear neural activation function  
and uses orthogonal reservoir matrices. Our numerical results have shown that the 
system's memory capacity is higher than the dimensionality of the reservoir, 
which is the upper bound for the typical RC approach based on ESNs. 
We have also discussed the application of the system to symmetric cryptography, 
and we made several suggestions to enhance its robustness and security. 
A numerical implementation was also included in the Appendix. 

\section*{Appendix}

Here we give a Python (Numpy) implementation for the case with cyclic 
permutations, which is also the fastest. The program learns the first 
paragraph from The Adventures of Sherlock Holmes by Sir Arthur Conan Doyle. 
The text has $T=1137$ characters, with $M=38$ distinct characters. 
With the parameters from the program, $N=T/2$ 
and $\alpha=0.5$, the online learning method converges in 291 iterations.

\begin{footnotesize}
\begin{verbatim}
# Reservoir computing on the hypersphere
import numpy as np

def init(M,N):
  u,v = np.random.rand(N,M),np.identity(M)
  for m in range(M):
    u[:,m] = u[:,m] - u[:,m].mean()
    u[:,m] = u[:,m]/np.linalg.norm(u[:,m])
  return u,v

def recall(T,N,w,u,c,a,ss):
  x,i = np.zeros(N),ci[ss]
  for t in range(T-1):
    x = (1.0-a)*x + a*(u[:,i] + np.roll(x,1))
    x = x/np.linalg.norm(x)
    y = np.exp(np.dot(w,x))
    i = np.argmax(y/np.sum(y))
    ss = ss + str(c[i])
  return ss

def error(s,ss):
  err = 0.
  for t in range(len(s)):
    err = err + (s[t]!=ss[t])
  return np.round(err*100.0/len(s),2)

def offline_learning(u,v,c,a,s):
  T,(N,M),eta = len(s),u.shape,1e-7
  X,S,x = np.zeros((N,T-1)),np.zeros((M,T-1)),np.zeros(N)
  for t in range(T-1):
    x = (1.0-a)*x + a*(u[:,ci[s[t]]] + np.roll(x,1))
    x = x/np.linalg.norm(x)
    X[:,t],S[:,t] = x,v[:,ci[s[t+1]]]
  XX = np.dot(X,X.T)
  for n in range(N):
    XX[n,n] = XX[n,n] + eta
  w = np.dot(np.dot(S,X.T),np.linalg.inv(XX))
  ss = recall(T,N,w,u,c,alpha,s[0])
  print "err=",error(s,ss),"%\n",ss,"\n"
  return ss,w

def online_learning(u,v,c,a,s):
  T,(N,M) = len(s),u.shape
  w,err,tt = np.zeros((M,N)),100.,0
  while err>0 and tt<T:
    x = np.zeros(N)
    for t in range(T-1):
      x = (1.0-a)*x + a*(u[:,ci[s[t]]] + np.roll(x,1))
      x = x/np.linalg.norm(x)
      p = np.exp(np.dot(w,x))
      p = p/np.sum(p)
      w = w + np.outer(v[:,ci[s[t+1]]]-p,x)
    ss = recall(T,N,w,u,c,a,s[0])
    err,tt = error(s,ss),tt+1
    print tt,"err=",err,"%\n",ss,"\n"
  return ss,w

s = \
"To Sherlock Holmes she is always THE woman. I have seldom heard him mention \
her any other name. In his eyes she eclipses and predominates the whole of \
her sex. It was not that he felt any emotion akin to love for Irene Adler. \
All emotions, and that one particularly, were abhorrent to his cold, precise \
but admirably balanced mind. He was, I take it, the most perfect reasoning \
and observing machine that the world has seen, but as a lover he would have \
placed himself in a false position. He never spoke of the softer passions, \
save with a gibe and a sneer. They were admirable things for the observer--\
excellent for drawing the veil from men's motives and actions. But for the \
trained reasoner to admit such intrusions into his own delicate and finely \
adjusted temperament was to introduce a distracting factor which might throw \
a doubt upon all his mental results. Grit in a sensitive instrument, or a crack \
in one of his own high-power lenses, would not be more disturbing than a strong \
emotion in a nature such as his. And yet there was but one woman to him, and \
that woman was the late Irene Adler, of dubious and questionable memory."

c = list(set(s))
ci = {ch:m for m,ch in enumerate(c)}
T,M = len(s),len(c)
N,alpha = int(0.5*T),0.5

np.random.seed(12345)
u,v = init(M,N)

ss,w = offline_learning(u,v,c,alpha,s)
ss,w = online_learning(u,v,c,alpha,s)

print T,N,M,alpha
\end{verbatim}
\end{footnotesize}


\begin{thebibliography}{00}

\bibitem{key-1} H. T. Siegelmann, E. D. Sontag, 
				{\it Applied Mathematics Letters}, {\bf 4}(6),77 (1991).
\bibitem{key-2} K. Funahashi, Y. Nakamura, 
				{\it Neural networks}, {\bf 6}(6), 801 (1993).
\bibitem{key-3} Y. Bengio, P. Simard, P. Frasconi, 
				{\it IEEE Trans. Neural Networks}, {\bf 5}, 157 (1994).				
\bibitem{key-4} M. Luko\v{s}evi\v{c}ius, H. Jaeger, 
				{\it Computer Science Review}, {\bf 3}(3),127 (2009).
\bibitem{key-5} H. Jaeger, 
				The "echo state" approach to analysing and training recurrent neural networks, 
				Technical Report GMD Report 148, German National Research Center for Information Technology (2001).
\bibitem{key-6} M. Luko\v{s}evi\v{c}ius, H. Jaeger, 
				Overview of reservoir recipes, School Eng. Sci., Jacobs Univ., Bremen, Germany, Tech. Rep. 11 (2007).	
\bibitem{key-7} M. Luko\v{s}evi\v{c}ius, 
				A Practical Guide to Applying Echo State Networks,
				Neural Networks: Tricks of the Trade, Lecture Notes in Computer Science 7700, 659 (2012).
\bibitem{key-8} H. Jaeger, 
				Short term memory in echo state networks. Technical Report GMD Report 152, 
				German National Research Center for Information Technology (2001).
\bibitem{key-9} T. Strau{\ss}, W. Wustlich, R. Labahn,
				{\it Neural Computation}, {\bf 24}(12), 3246 (2012).
\bibitem{key-10} R. Ramamurthy, C. Bauckhage, K. Buza, S. Wrobel,
				arXiv:1704.01046 (2017).
\end{thebibliography}
\end{document}